\def\year{2018}\relax
\documentclass[letterpaper]{article} 
\usepackage{aaai18}  
\usepackage{times}  
\usepackage{helvet}  
\usepackage{courier}  
\usepackage{url}  
\usepackage{graphicx}  

\usepackage{bm}
\usepackage{amsfonts}
\usepackage{amsmath}

\usepackage{color}
\usepackage{subfigure}
\usepackage{booktabs}
\usepackage{multirow}
\usepackage{array}

\begin{document}
%
\title{SEE: Syntax-aware Entity Embedding for Neural Relation Extraction}
\author{Zhengqiu He\textsuperscript{1},
Wenliang Chen\textsuperscript{1,4}\thanks{The corresponding author is Wenliang Chen.},
Zhenghua Li\textsuperscript{1} \\
{ \bf \Large Meishan Zhang\textsuperscript{3},
Wei Zhang\textsuperscript{2},
Min Zhang\textsuperscript{1}} \\
\textsuperscript{1}School of Computer Science and Technology, Soochow
University, China\\
\textsuperscript{2}Alibaba Group, China\\
\textsuperscript{3}School of Computer Science and Technology, Heilongjiang
University, China\\
\textsuperscript{4}Collaborative Innovation Center of Novel Software Technology and Industrialization, China\\
zqhe@stu.suda.edu.cn,
\{wlchen, zhli13, minzhang\}@suda.edu.cn\\
mason.zms@gmail.com,
lantu.zw@alibaba-inc.com}

\maketitle
\begin{abstract}
\begin{quote}

Distant supervised relation extraction is an efficient approach
to scale relation extraction to very large corpora,
and has been widely used to find novel relational
facts from plain text. Recent studies on neural relation extraction have shown great progress on this task via modeling the sentences
in low-dimensional spaces, but seldom considered syntax information to model the entities.
In this paper, we propose to learn syntax-aware entity embedding for neural relation extraction.
First, we encode the context of entities on a dependency tree as sentence-level entity embedding based on tree-GRU.
Then, we utilize both intra-sentence
and inter-sentence attentions to obtain sentence set-level entity embedding over all sentences containing the focus entity pair.
Finally, we combine both sentence embedding and entity embedding for relation classification.
We conduct experiments on a widely used real-world
dataset and the experimental results show that
our model can make full use of all informative instances
and achieve state-of-the-art performance of relation extraction.
\end{quote}
\end{abstract}

\section{Introduction}
Relation extraction (RE), defined as the task of extracting semantic relations between entity pairs from plain text, has received increasing interests in the community of natural language processing \cite{riedel2013relation,miwa2016endtoend}. The task is a typical classification problem after the entity pairs are specified \cite{Zeng2014Relation}. Traditional supervised methods require large-scale manually-constructed corpus, which is expensive and confined to certain domains. Recently, distant supervision has gained a lot of attentions which is capable of exploiting automatically-produced training corpus \cite{mintz2009distant}. The framework has achieved  great success and has brought state-of-the-art performances in RE.

Given an entity pair ($e',\, e''$) from one knowledge base (KB) such as Freebase, assuming that the predefined semantic relation on the KB is $r$, we simply label all sentences containing the two entities by label $r$. This is the key principle for distant supervision to produce training corpus. While this may be problematic in some conditions, thus can result in noises. For example, the sentence ``\emph{Investors include Vinod Khosla of Khosla Ventures, who, with the private equity group of texas pacific group ventures, invested \$20 million.}" is not for relation \emph{/business/company/founders} of \emph{Khosla Ventures} and \emph{Vinod Khosla} in Freebase, but it is still be regarded as a positive instance under the assumption of distant supervision. Based on the observation, recent work present multi-instance learning (MIL) to address the problem, by treating each produced sentence differently during the training \cite{riedel2010modeling,zeng2015distant,Lin2016Neural}.
Our work also falls into this category.
\begin{figure}[tb]
  \centering
\begin{minipage}{0.45\textwidth}
\centering
          \includegraphics[width=0.9\textwidth]{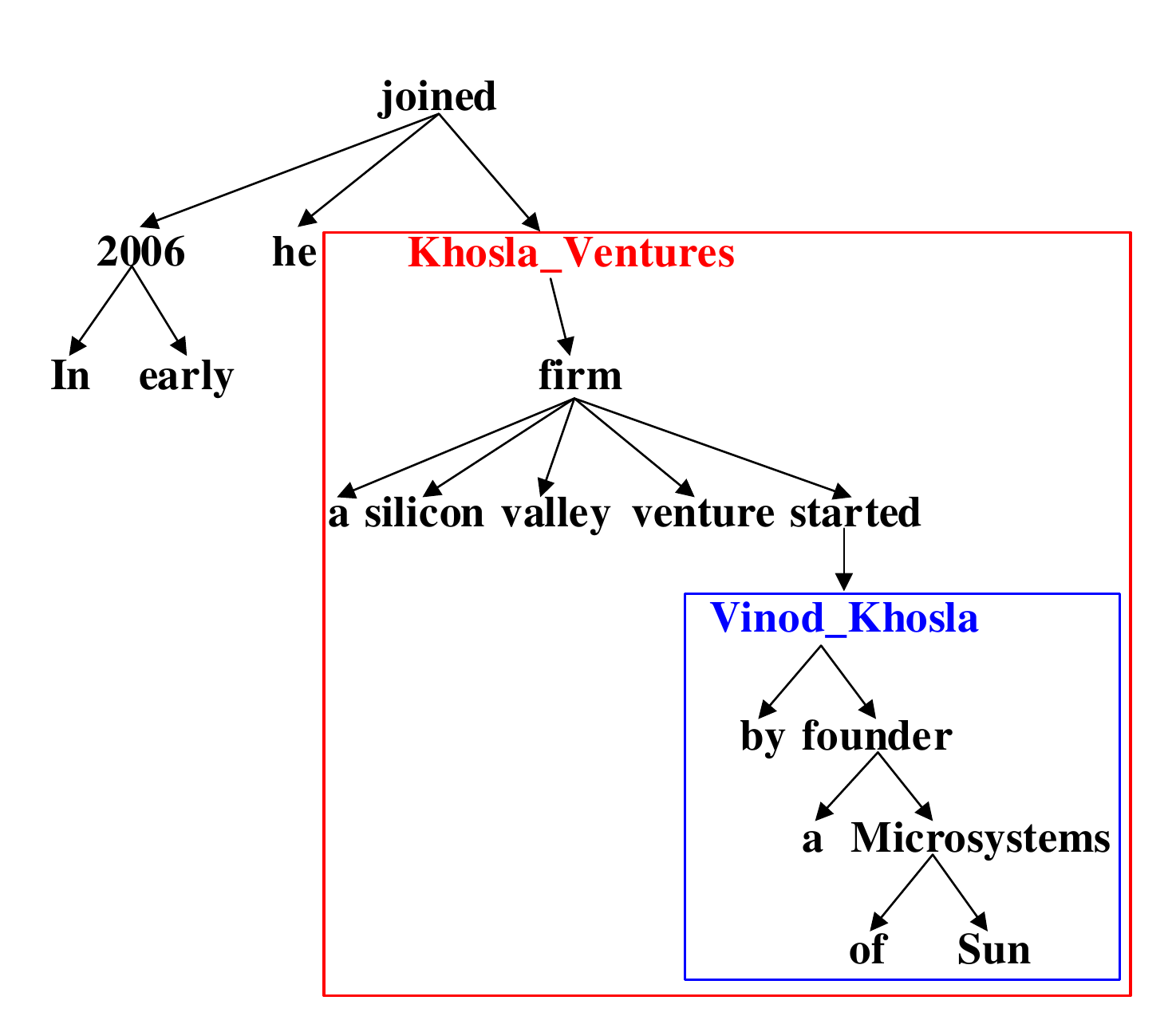}
      \end{minipage}
  \caption{An example dependency tree containing two entities in sentence ``In early 2006, he joined Khosla Ventures, a silicon valley venture firm started by Vinod Khosla, a founder of Sun Microsystems.".}\label{entity}
\end{figure}

Under the statistical models with handcrafted features, a number of studies have proposed syntactic features, and achieved better results by using them \cite{hoffmann2011knowledge,surdeanu2012multi}. Recently, the neural network models have dominated the work of RE because of higher performances \cite{Lin2016Neural,ji2017distant}. Similarly, the syntax information has also been investigated in neural RE. One representative method is to use the shortest dependency path (SDP) between a given entity pair \cite{miwa2016endtoend}, based on which long short term memory (LSTM) can be applied naturally to model it. This method has brought remarkable results, since the path words are indeed good indictors for semantic relation and meanwhile SDPs can remove abundant words between entity pairs.

The above work of using syntax concerns mainly on the connections between entity pairs, paying much attention on the words that link the two entities semantically, while neglects the representation of entities themselves. Previous entity embeddings purely based on their sequential words can be insufficient to generalize to unknown entities. But it can be different when we try to capture the meaning of entities by its syntactic contexts. For example, as shown in Figure \ref{entity}, when use the subtrees rooted at \emph{Khosla Ventures} and \emph{Vinod Khosla} to represent the two entities, we could capture longer distance information than only use the entities themselves. It indicates that the syntax roles the entities played in the sentences are informative for RE.

\begin{figure}[tb]
\centering
\begin{minipage}{0.55\textwidth}
          \includegraphics[width=0.88\textwidth]{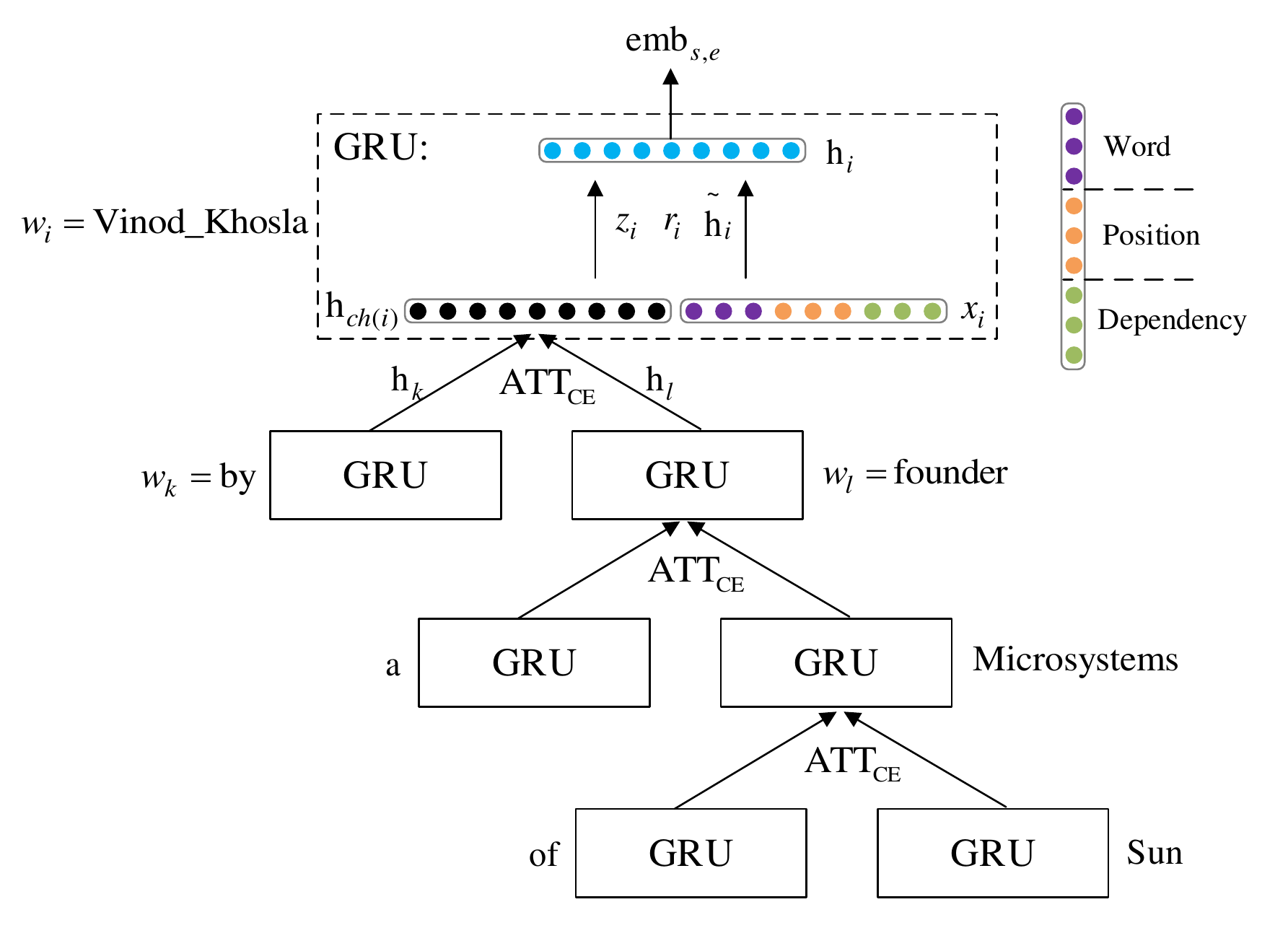}
      \end{minipage}
  \caption{Workflow of entity embedding via tree-GRU.}\label{context}
\end{figure}

In this paper, we propose syntax-aware entity embedding (SEE) for enhancing neural relation extraction. As illustrated in Figure \ref{context}, to enrich the representation of each entity, we build tree-structured recursive neural networks with gated recursive units (tree-GRU) to embed the semantics of entity contexts on dependency trees. Moreover, we employ both intra-sentence and inter-sentence attentions to make full use of syntactic contexts in all sentences: (1) attention over child embeddings in a parse tree to distinguish informative children; (2) attention over sentence-level entity embeddings to alleviate the wrong label problem.
Finally, we combine all sentence embeddings and entity embeddings for relation classification. We evaluate our model on the widely used benchmark dataset and show that our proposed model achieves consistently better performance than the state-of-the-art methods.

\section{The Baseline}

Our baseline model directly adopts the state-of-the-art neural relation extraction model proposed by \citeauthor{Lin2016Neural} \shortcite{Lin2016Neural}, which also employs multi-instance learning for alleviating the wrong label problem faced by the distant supervision paradigm.

\begin{figure}[tb]
\centering
\begin{minipage}{0.5\textwidth}
      \centering
          \includegraphics[width=0.88\textwidth]{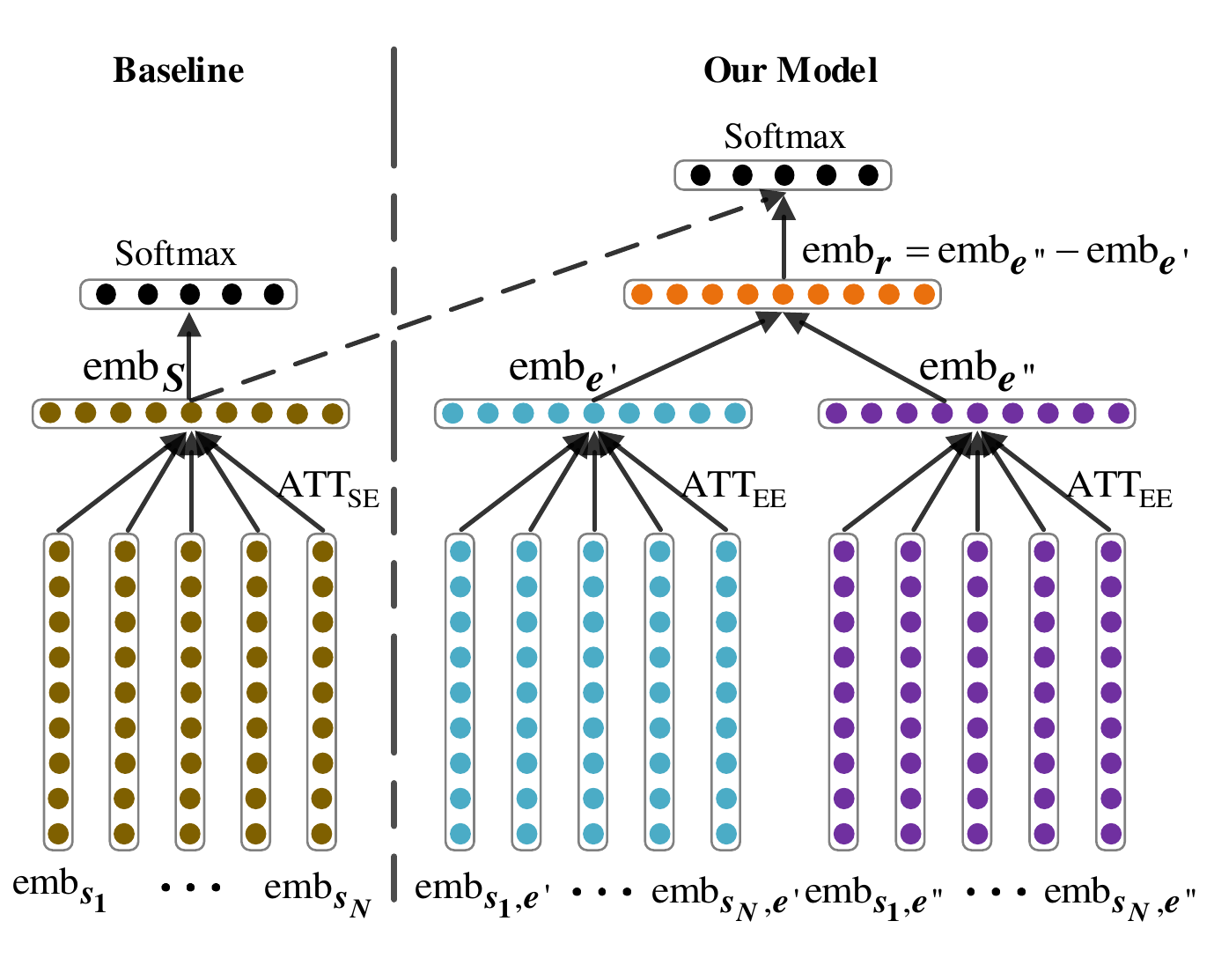}
      \end{minipage}
  \caption{Workflow of the baseline and our approach.}\label{the model}
\end{figure}

The framework of the baseline approach is illustrated in the left part of Figure \ref{the model}.
Suppose there are $N$ sentences $S
= \{s_1, ..., s_N\}$ that contain the focus entity pair $e'$ and $e''$.
The input is the embeddings of all the sentences.
The $i$-th sentence embedding, i.e., $\mathbf{emb}_{s_i}$, is built from the word sequence, and encodes the semantic representation of the corresponding sentence.
Then, an attention layer is performed to obtain the representation vector of the sentence set.
Finally, a softmax layer produces the probabilities of all relation types.

\subsection{Sentence Embedding}

\begin{figure}[tb]
  \centering
\begin{minipage}{0.45\textwidth}
          \includegraphics[width=0.83\textwidth]{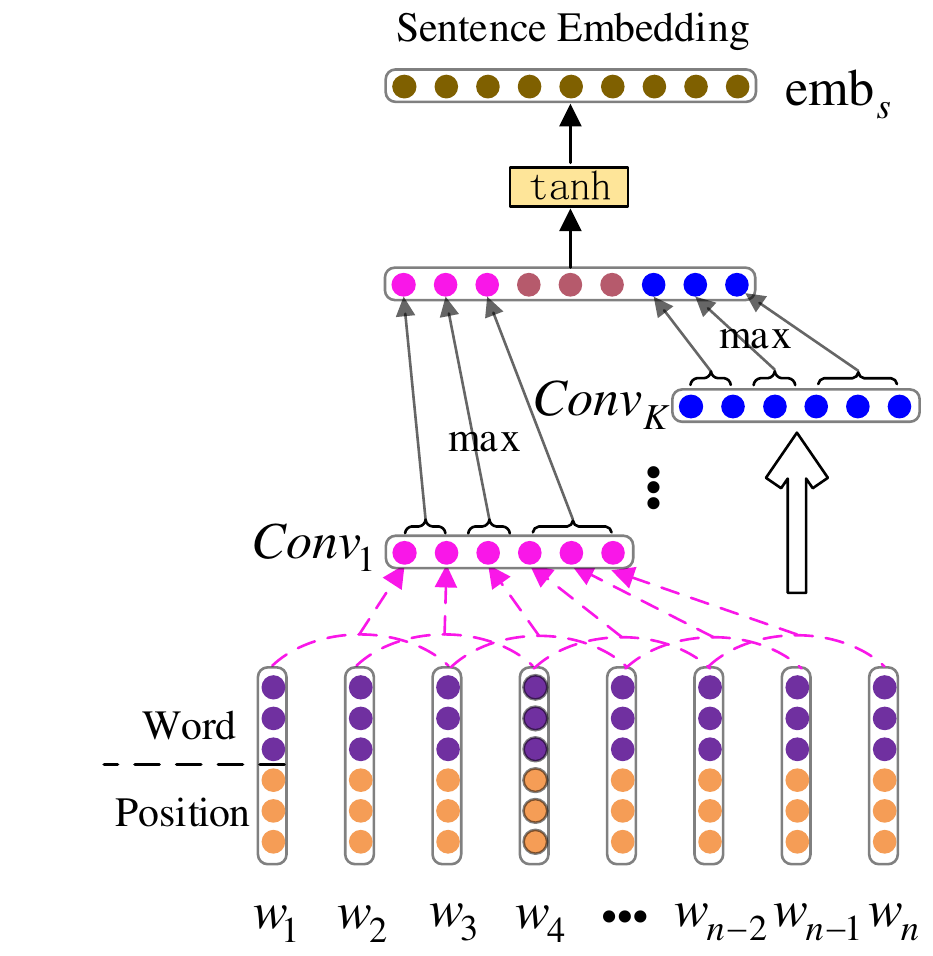}
      \end{minipage}
  \caption{Workflow of sentence embedding.}\label{Sentence Vector}
\end{figure}

Figure \ref{Sentence Vector} describes the component for building a sentence embedding from the word sequence.
Given a sentence $s=\{w_1, ..., w_n\}$, where $w_i$ is the $i$-th word in the sentence,
the input is a matrix composed of $n$ vectors $\mathbf{X} = [\bm{x}_1, ..., \bm{x}_n]$,
where $\bm{x}_i$ corresponds to $w_i$ and consists of the word embedding
and its position embedding.
Following \citeauthor{zeng2015distant} \shortcite{zeng2015distant} and \citeauthor{Lin2016Neural} \shortcite{Lin2016Neural}, we employ the skip-gram method of \citeauthor{mikolov2013b} \shortcite{mikolov2013b} to pretrain the word embeddings, which will be fine-tuned afterwards.
Position embeddings are first successfully applied to relation extraction by \citeauthor{Zeng2014Relation} \shortcite{Zeng2014Relation}.
Given a word (e.g., ``firm'' in Figure \ref{entity}), its position embedding corresponds to the relative distance (``6\&-3'') from the word to the entity pairs (``Khosla Ventures'' and ``Vinod Khosla'') through lookup.

\textbf{A convolution layer} is then applied to reconstruct the original input $\mathbf{X}$ by learning sentence features from a small window of words at a time while preserving word order information.
They use $K$ convolution filters (a.k.a. feature maps) with the same window size $l$.
The $j$-th filter uses a weight matrix $\mathbf{W}^f_j$ to map $\bm{X}$ into a $j$-th-view vector $\mathbf{Conv}_j(\bm{X})$, which contains $n-l+1$ scalar elements. The $i$-th element is computed as follows:
$$\mathbf{Conv}_j(\bm{X})[i]=\mathbf{W}^f_j ~ \bm{X}_{i:i+l-1}$$

\textbf{Three-segment max-pooling} is then applied to map $K$ convolution output vectors of varying length into a vector of a fixed length $3K$.
Suppose the positions of the two entities are $p_1$ and $p_2$ respectively.\footnote{\citeauthor{Lin2016Neural} \shortcite{Lin2016Neural} treat all entity names as single words.}
Then, each convolution output vector $\mathbf{Conv}_j(\bm{X})$ is divided into three segments:
$$[0:p_1-1]/[p_1:p_2]/[p_2+1:n-l]$$

The max scalars in each segment is preserved to form a $3$-element vector, and all vectors produced by the $K$ filters are concatenated into a $3K$-element vector, which is the output of the pooling layer.\footnote{The combination of CNN and three-segment Max-pooling is first proposed by \citeauthor{zeng2015distant} \shortcite{zeng2015distant} and named as piecewise convolutional neural network (PCNN).}

Finally, the sentence embedding $\mathbf{emb}_s$ is obtained after a non-linear  transformation (e.g., tanh) on the $3K$-element vector.

\subsection{Relation Classification}

\textbf{An attention layer over sentence embeddings (ATT$_\textit{SE}$)} is performed over the input sentence embeddings ($\mathbf{emb}_{s_i}, 1 \le i \le N$) to produce a vector that encodes the sentence set, 
as shown in Figure \ref{the model}. We adopt the recently proposed self-attention method \cite{Lin2017A}. First, each
sentence $s_i$ gains an attention score as follows:
$$ \alpha_i = \mathbf{v}^\textit{sa} \, \mathrm{tanh}(\mathbf{W}^\textit{sa} \mathbf{emb}_{s_i})$$

where the matrix $\mathbf{W}^\textit{sa}$ and the vector $\mathbf{v}^\textit{sa}$ are the sentence attention parameters.

Then, the attention scores are normalized into a probability for summing all sentence embeddings into the representation vector of the sentence set $S$.
As discussed in \citeauthor{Lin2016Neural} \shortcite{Lin2016Neural}, the attention layer aims to automatically detect noisy training sentences with wrong labels by allocating lower weights to them in this step.\footnote{Please note that \citeauthor{Lin2016Neural} \shortcite{Lin2016Neural} actually use a more complicated attention schema. However, our preliminary experiments show that the simple self-attention method presented here can achieve nearly the same accuracy. Moreover, the same self-attention mechanism is employed as both local and global attention in our proposed approach. }

\begin{equation}\label{eq:att-ove-sentence}
\mathbf{emb}_S = \sum_{1 \le i \le N} \left\{
\frac{ \mathrm{exp}(\alpha_i) }{ \sum\limits_{1 \le k \le N}{ \mathrm{exp}(\alpha_k) } }
\mathbf{emb}_{s_i}
\right\}
\end{equation}

\textbf{A softmax layer} is used to
produce the probabilities of all relation types.
First, we compute a output score vector as follows:
\begin{equation}\label{eq:output-rel-score}
\mathbf{o}^\textit{s} = \mathbf{W}^\textit{s} \mathbf{emb}_S  + \mathbf{b}^\textit{s}
\end{equation}
where the matrix $\mathbf{W}^\textit{s}$ and the bias vector $\mathbf{b}^\textit{s}$ are model parameters, and
$|\mathbf{o}^\textit{s}|=N_r$ is the number of relation types.

Then, the conditional probability of the relation $r$ for given $S$ is:
\begin{equation}\label{eq:softmax-rel-prob}
p(r|S) = \frac{\mathrm{exp}(\mathbf{o}^\textit{s}[r])}
{\sum_{1 \le k \le N_r} {\mathrm{exp}(\mathbf{o}^\textit{s}[k])}}
\end{equation}

\subsection{Training Objective}

Given the training data $\mathcal{D} = \{(S_1,r_1), ..., (S_M, r_M)\}$ consisting of $M$ sentence sets and their relation types resulting from distant supervision,
\citeauthor{Lin2016Neural} \shortcite{Lin2016Neural} use the standard cross-entropy loss function as the training objective.
\begin{equation}\label{eq:cross-entropy-loss}
Loss(\mathcal{D}) = - \sum_{i=1}^{M}\mathrm{log}\,p(r_i|S_i)
\end{equation}
Following \citeauthor{Lin2016Neural} \shortcite{Lin2016Neural}, we adopt stochastic gradient descent (SGD) with mini-batch as the learning algorithm and apply dropout \cite{srivastava2014dropout} in Equation (\ref{eq:output-rel-score}) to prevent over-fitting.

\section{Our SEE Approach}

The baseline approach solely relies on the word sequence of a given sentence.
However, recent studies show that syntactic structures can help relation extraction by exploiting the dependence relationship between words.
Unlike previous works which mainly consider the shortest dependency paths, our proposed approach tries
to effectively encode the syntax-aware contexts of entities as extra features for relation classification.

\subsection{Entity Embedding}

Given a sentence and its parse tree, as depicted in Figure \ref{entity}, we
try to encode the 
focus entity pair as two dense vectors.

Previous work shows that recursive neural networks (RNN) are effective in encoding tree structures \cite{Li2015When}.
Inspired by \citeauthor{tai2015improved} \shortcite{tai2015improved},
we propose a simple attention-based tree-GRU to
derive the context embedding of an entity over its dependency subtree in a bottom-up order.\footnote{In fact, \citeauthor{tai2015improved} \shortcite{tai2015improved} propose two extensions to the basic LSTM architecture, i.e., the \emph{N-ary tree-LSTM} and the \emph{child-sum tree-LSTM}. However, the \emph{N-ary tree-LSTM} assumes that the maximum number of children is $N$, which may be unsuitable for our task since $N=19$ would be too large for our dataset.
The \emph{child-sum tree-LSTM} can handle arbitrary number of children, but
achieves consistently lower accuracy than the simple attention-based tree-GRU according to our preliminary experiments.
}

Figure \ref{context} illustrates the attention-based tree-GRU.
Each word corresponds to a GRU node.
Suppose ``Vinod\_Khosla'' is the $i$-th word $w_i$ in the sentence, and take its corresponding GRU node as an example.
The GRU node has two input vectors.
The first input vector, denoted as $\bm{x}_i$, consists of the word embedding, the position embedding, and the dependency embedding of ``started   $\rightarrow$ Vinod\_Khosla''. It is similar to the input in Figure \ref{Sentence Vector} except for the extra dependency embedding. 


\textbf{A dependency embedding} is a dense vector that encodes a head-modifier word pair in contexts of all dependency trees, which can express richer semantic relationships beyond word embedding, especially for long-distance collocations.
Inspired by \citeauthor{bansal2015dependency} \shortcite{bansal2015dependency}, we adopt the skip-gram neural language model of \citeauthor{mikolov2013a} \shortcite{mikolov2013a,mikolov2013b} to learn the dependency embedding.
First, we employ the off-shelf Stanford Parser\footnote{https://nlp.stanford.edu/software/lex-parser.shtml, and the version is 3.7.0} to parse the New York Times (NYT) corpus  \cite{Klein2003Accurate}.
Then, given a father-child dependency $p \rightarrow c$, the skip-gram model is optimized to predict all its context dependencies.
We use the following basic dependencies in a parse tree as contexts:
$$gp \rightarrow p \quad c \rightarrow  gc_{1} \quad \dots \quad c \rightarrow gc_{\#gc}$$
where $gp$ means grandparent; $gc$ means grandchild; $\#gc$ is the total number of grandchildren.

The second input vector of the GRU node of ``Vinod\_Khosla'' is the representation vector of all its children $\mathbf{ch}(i)$, and is denoted as $\bm{h}_{\mathbf{ch}(i)}$.

\textbf{Attention over child embeddings (ATT$_\textit{CE}$)}.
Here, we adopt the self-attention for summing the hidden vector of the GRU nodes of its children.
Suppose $j \in \mathbf{ch}(i)$, meaning $w_j$ is a child of $w_i$.
We use  $\bm{h}_{j}$ to represent the hidden vector of the GRU node of $w_j$.
Then, the attention score of $\bm{h}_{j}$ is:
\begin{equation*}\label{eq:attention-over-children}
\alpha^i_j = \mathbf{v}^\textit{ch} \, \mathrm{tanh}(\mathbf{W}^\textit{ch}  \bm{h}_{j})
\end{equation*}
where $\mathbf{v}^\textit{ch}$ and $\mathbf{W}^\textit{ch}$ are shared attention parameters.

Then, the children representation vector is computed as:
\begin{equation}\label{eq:children-embedding}
\bm{h}_{\mathbf{ch}(i)} = \sum_{j \in \mathbf{ch}(i)} \left\{
\frac{ \mathrm{exp}(\alpha^i_j) }{ \sum\limits_{k \in \mathbf{ch}(i)}{ \mathrm{exp}(\alpha^i_k) } }
\mathbf{h}_{j}
\right\}
\end{equation}
We expect that the \textbf{ATT$_\textit{CE}$} mechanism can be helpful for producing better representation of the father by
1) automatically detecting informative children via higher attention weights; 2) whereas lowering the weights of incorrect dependencies due to parsing errors.

Given the two input vectors $\bm{x}_i$ and $\bm{h}_{\mathbf{ch}(i)}$,
the GRU node \cite{cho-al-emnlp14} computes the hidden vector of $w_i$ as follows:
\begin{equation}\label{tree_gru}
\begin{split}
  & \bm{z}_i = \sigma(\mathbf{W}^{z}\bm{x}_i + \mathbf{U}^{z}\bm{h}_{\mathbf{ch}(i)} + \mathbf{b}^{z})  \\
  & \bm{r}_i = \sigma(\mathbf{W}^{r}\bm{x}_i + \mathbf{U}^{r}\bm{h}_{\mathbf{ch}(i)} + \mathbf{b}^{r})  \\
  & \widetilde{\bm{h}}_i = \mathrm{tanh}(\mathbf{W}^{\widetilde{h}}\bm{x}_i + \mathbf{U}^{\widetilde{h}}(\bm{r}_i \circ \bm{h}_{\mathbf{ch}(i)}) + \mathbf{b}^{\widetilde{h}})  \\
  & \bm{h}_i = \bm{z}_i \circ \bm{h}_{\mathbf{ch}(i)} + (1-\bm{z}_i) \circ \widetilde{\bm{h}}_i  \\
\end{split}
\end{equation}
\noindent where $\sigma$ is the sigmoid function, and the $\circ$ is the element-wise multiplication, $\mathbf{W}^*$ and $\mathbf{U}^*$ are parameter matrices of the model, $\mathbf{b}^*$ is the bias vectors, $\bm{z}_i$ is the update gate vector and $\bm{r}_i$ is the reset gate vector.

Finally, we use $\bm{h}_i$ as the representation vector of the entity context of ``Vinod\_Khosla''.
In the same manner, we can compute the entity context embedding of ``Khosla\_Ventures''.

\subsection{Augmented Relation Classification}

Again, we suppose there are $N$ sentences $S =\{s_1, ..., s_N\}$ that contain the focus entity pair $e'$ and $e''$.
The corresponding word indices that $e'$ occurs in $S$ are respectively $\{j'_1, ..., j'_N\}$, whereas
the positions of $e''$ are $\{j''_1, ..., j''_N\}$.

As discussed above, the entity context embedding of $e'$ in the $i$-th sentence $s_i$ is the hidden vector of the GRU node of $w_{j'_i}$ (which is $e'$).
$$ \mathbf{emb}_{s_i, e'} = \bm{h}^{s_i}_{j'_i}$$

Similarly, the entity context embedding of $e''$ in $s_i$ is:
$$ \mathbf{emb}_{s_i, e''} = \bm{h}^{s_i}_{j''_i}$$

Figure \ref{the model} shows the overall framework of our proposed approach.
The input consists of three parts, i.e., the sentence embeddings , the context embeddings of $e'$, and the context embeddings of $e''$:
\begin{equation}\label{eq:three-embeddings}
\begin{split}
\{\mathbf{emb}_{s_1} ~~~ &  ... ~~~\mathbf{emb}_{s_N}\} \\
\{\mathbf{emb}_{s_1, e'} ~~~& ... ~~~\mathbf{emb}_{s_N,e'}\} \\
\{\mathbf{emb}_{s_1, e''} ~~~& ... ~~~\mathbf{emb}_{s_N,e''}\} \\
\end{split}
\end{equation}

Similar to sentence attention in the baseline system, and for maximizing utilization the valid information in sentence and entity context, we enhance the model by separately applying
attention to both the sentence and entity context embeddings simultaneously.

\textbf{Attention over entity embeddings (ATT$_\textit{EE}$)}. Similar to the attention over sentence embeddings in Equation (\ref{eq:att-ove-sentence}), we separately apply attention to the three parts in Equation (\ref{eq:three-embeddings}) and generate the final representation vectors of $S$, $e'$, and $e''$ on the sentence set, i.e., $\mathbf{emb}_S$, $\mathbf{emb}_{e'}$,  $\mathbf{emb}_{e''}$, respectively.
We omit the formulas for brevity.

Then, the next step is to predict the relation type based on the three sentence set-level embeddings. Here, we propose two strategies.

\textbf{The concatenation strategy (CAT)}. The most straightforward way is to directly concatenate the three embeddings and obtain the score vector of all relation types via a linear transformation.
\begin{equation} \label{eq:scores-cat}
\mathbf{o}^\textit{cat} = \mathbf{W}^\textit{cat}[\mathbf{emb}_{S};\mathbf{emb}_{e'};\mathbf{emb}_{e''}] + \mathbf{b}^\textit{cat}
\end{equation}
where the matrix $\mathbf{W}^\textit{cat}$ and the bias vector $\mathbf{b}^\textit{cat}$ are model parameters.

\textbf{The translation strategy (TRANS)}. According to Equation (\ref{eq:scores-cat}), the CAT strategy cannot capture the interactions among the three embeddings, which is counter-intuitive considering that the relation type must be closely related with both entities simultaneously.
Inspired by the widely used TransE model \cite{bordes2013translating}, which regards the embedding of a relation type $r$ as the difference between two entity embeddings ($\mathbf{emb}_{r} = \mathbf{emb}_{e''} - \mathbf{emb}_{e'}$), we use the vector difference to produce a relation score vector via a linear transformation.
\begin{equation} \label{eq:scores-entity-context}
\mathbf{o}^\textit{see} = \mathbf{W}^\textit{see}(\mathbf{emb}_{e''} - \mathbf{emb}_{e'}) + \mathbf{b}^\textit{see}
\end{equation}
where $\mathbf{o}^\textit{see}$ represents the score vector according to the entity context embeddings, and
the matrix $\mathbf{W}^\textit{see}$ and the bias vector $\mathbf{b}^\textit{see}$ are model parameters.

To further utilize the sentence embeddings, we compute another relation score vector $\mathbf{o}^\textit{s}$ according to Equation (\ref{eq:output-rel-score}), which is the same with the baseline.
Then we combine the two score vectors.
\begin{equation}\label{eq:interpolation-score}
\begin{split}
  & \mathbf{o}^\textit{trans} = \bm{\alpha} \circ \mathbf{o}^\textit{s} + (1-\bm{\alpha})\circ \mathbf{o}^\textit{see} \\
\end{split}
\end{equation}
where $\circ$ denotes element-wise product (a.k.a. Hadamard product), and
$\bm{\alpha}$ is the interpolation vector for balancing the two parts.
Actually, we have also tried a few different ways for combining the two score vectors, but found that the formula presented here consistently performs best.

Finally, we apply softmax to transform the score vectors ($\mathbf{o}^\textit{cat}$ or $\mathbf{o}^\textit{trans}$) into conditional probabilities, as shown in Equation (\ref{eq:softmax-rel-prob}), and
 adopt the same training objective and optimization algorithm with the baseline.

\section{Experiments}

In this section, we present the experimental results and detailed analysis.

\textbf{Datasets.}
We adopt the benchmark dataset developed by \citeauthor{riedel2010modeling} \shortcite{riedel2010modeling}, which has been widely used in many recent works \cite{hoffmann2011knowledge,surdeanu2012multi,Lin2016Neural,ji2017distant}.
\citeauthor{riedel2010modeling} \shortcite{riedel2010modeling} use Freebase as the distant supervision source and the three-year NYT corpus from $2005$ to $2007$ as the text corpus.
First, they detect the entity names in the sentences using the Stanford named entity tagger \cite{finkel2005incorporating} for matching the Freebase entities.
Then, they project the entity-relation tuples in Freebase into the all sentences that contain the focus entity pair.
The dataset contains $53$ relation types, including a special relation ``NA" standing for no relation between the entity pair.
We adopt the standard data split (sentences in $2005$-$2006$ NYT data for training, and sentences in $2007$ for evaluation).
The training data contains $522,611$ sentences, $281,270$ entity pairs and $18,252$ relational facts. The testing set contains $172,448$ sentences, $96,678$ entity pairs and $1,950$ relational facts.

\begin{figure} [tb]
\begin{minipage}{0.48\textwidth}
      \centering
          \includegraphics[width=0.9\textwidth]{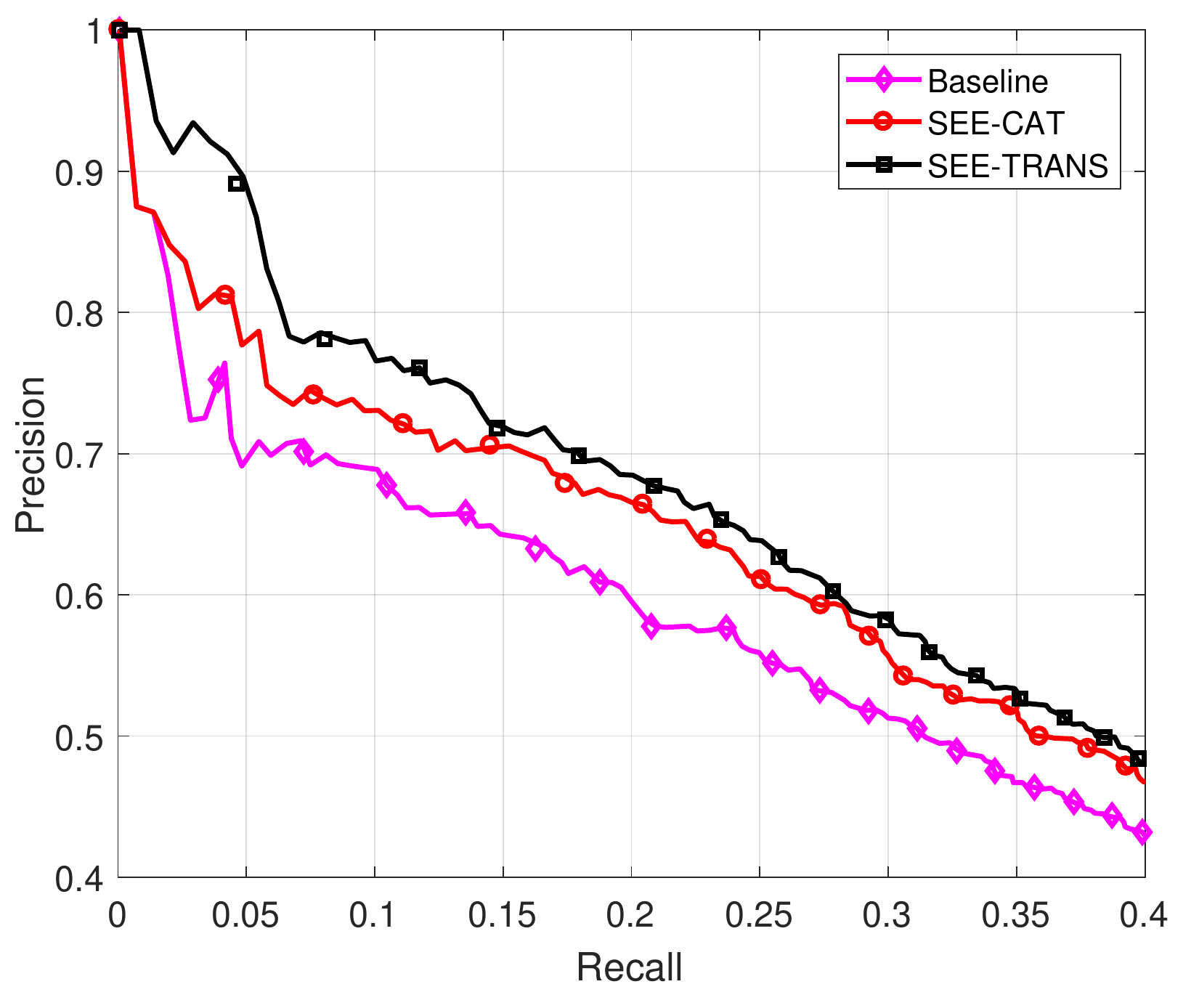}
      \end{minipage}
  \caption{Comparison of the baseline and our approach under two different strategies.}\label{baseline}
\end{figure}

\textbf{Evaluation metrics.}
Following the practice of previous works \cite{riedel2010modeling,zeng2015distant,ji2017distant}, we employ two evaluation methods, i.e., the \emph{held-out evaluation} and the \emph{manual evaluation}. The held-out evaluation only compares the entity-relation tuples produced by the system on the test data against the existing Freebase entity-relation tuples,   
and report the precision-recall curves.

Manual evaluation is performed to avoid the influence of the wrong labels resulting from distant supervision and the incompleteness of Freebase data, and report the Top-$N$ precision $P$@$N$, meaning the the precision of the top $N$ discovered relational facts with the highest probabilities.

\textbf{Hyperparameter tuning. }
We tune the hyper-parameters of all the baseline and our proposed models on the training dataset using three-fold validation.
We adopt the brute-force grid search to decide the optimal hyperparameters for each model. We try $\{0.1, 0.15, 0.2, 0.25\}$ for the initial learning rate of SGD,
$\{50, 100, 150, 200\}$ for the mini-batch size of SGD,
$\{50, 80, 100\}$ for both the word and the dependency embedding dimensions,
$\{5, 10, 20\}$ for the position embedding dimension,
$\{3, 5, 7\}$ for the convolution window size $l$,
and $\{60, 120, 180, 240, 300\}$ for the filter number $K$.
We find the configuration $0.2/150/50/50/5/3/240$ works well for all the models, and further tuning leads to slight improvement.

\subsection{Held-out Evaluation}

\textbf{Comparison results with the baseline} is presented in Figure \ref{baseline}.
``SEE-CAT'' and ``SEE-TRANS'' are our proposed approach with the CAT and TRANS strategies respectively.
We can see that both our approaches consistently outperform the baseline method.
It is also clear that ``SEE-TRANS'' is superior to ``SEE-CAT''.
This is consistent with our intuition that the TRANS strategy can better capture the interaction between the two entities simultaneously.
In the following results, we adopt ``SEE-TRANS'' for further experiments and analysis.

\begin{figure}[tb]
\begin{minipage}{0.48\textwidth}
      \centering
          \includegraphics[width=0.9\textwidth]{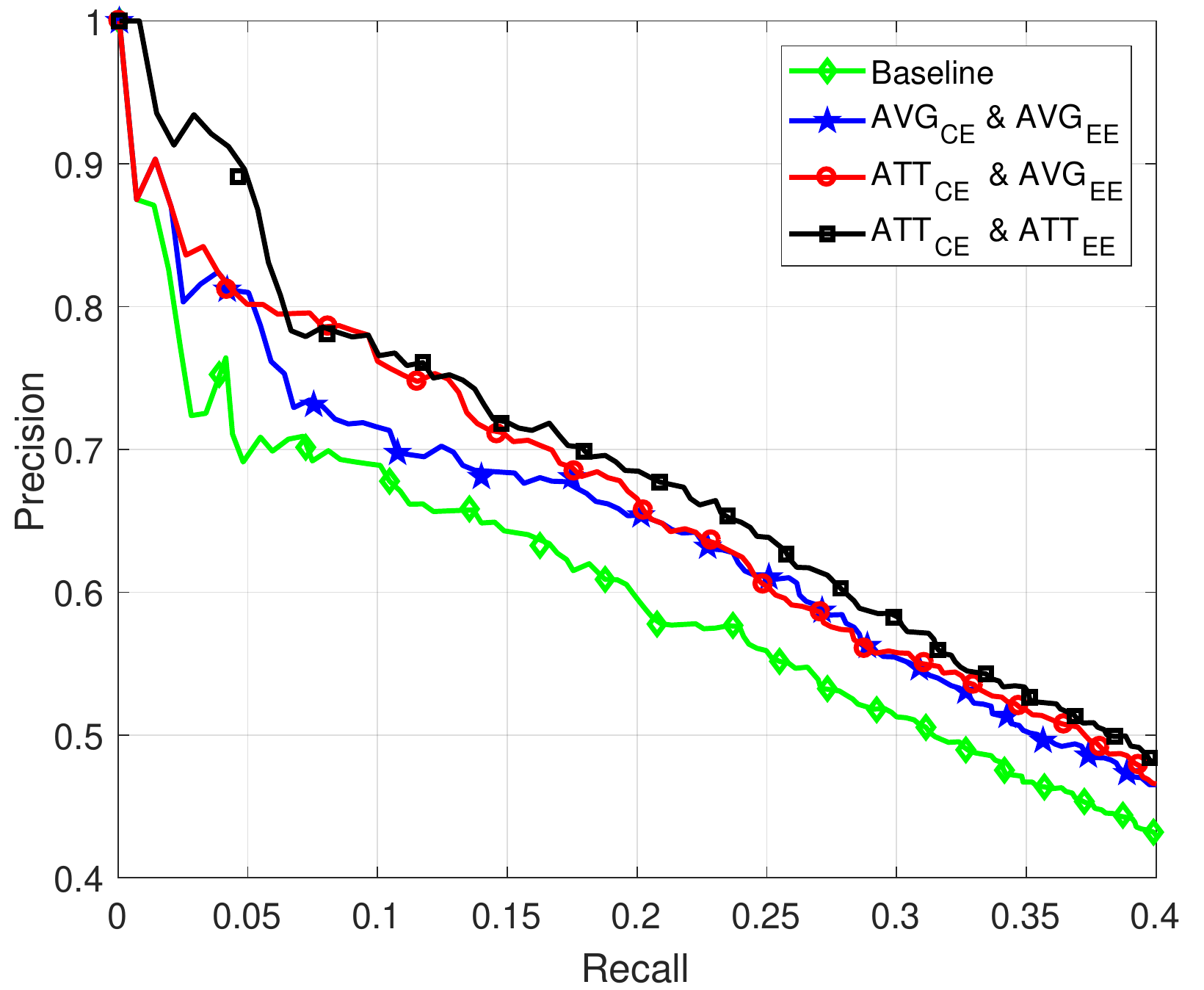}
      \end{minipage}

  \caption{Effect of self-attention components.}\label{compare}
\end{figure}

\textbf{The effect of self-attention components} is investigated in Figure \ref{compare}.
To better understand the two self-attention components used in our ``SEE'' approach,
we replace attention with an average component, which assumes the same weight for all input vectors and simply use the averaged vector as the resulting embedding.
Therefore, the ``ATT$_\mathrm{{CE}}$'' in Figure \ref{context} is replaced with  ``AVG$_\mathrm{{CE}}$'',
and ``ATT$_\mathrm{{EE}}$'' in Figure \ref{the model} is replaced with ``AVG$_\mathrm{{EE}}$''.

The four precision-recall curves clearly show that both self-attention components are helpful for our model.
In other words, the attention provides a flexible mechanism that allows the model to distinguish the contribution of different input vectors, leading to better global representation of instances.

\begin{figure}[tb]
\begin{minipage}{0.48\textwidth}
      \centering
          \includegraphics[width=0.9\textwidth]{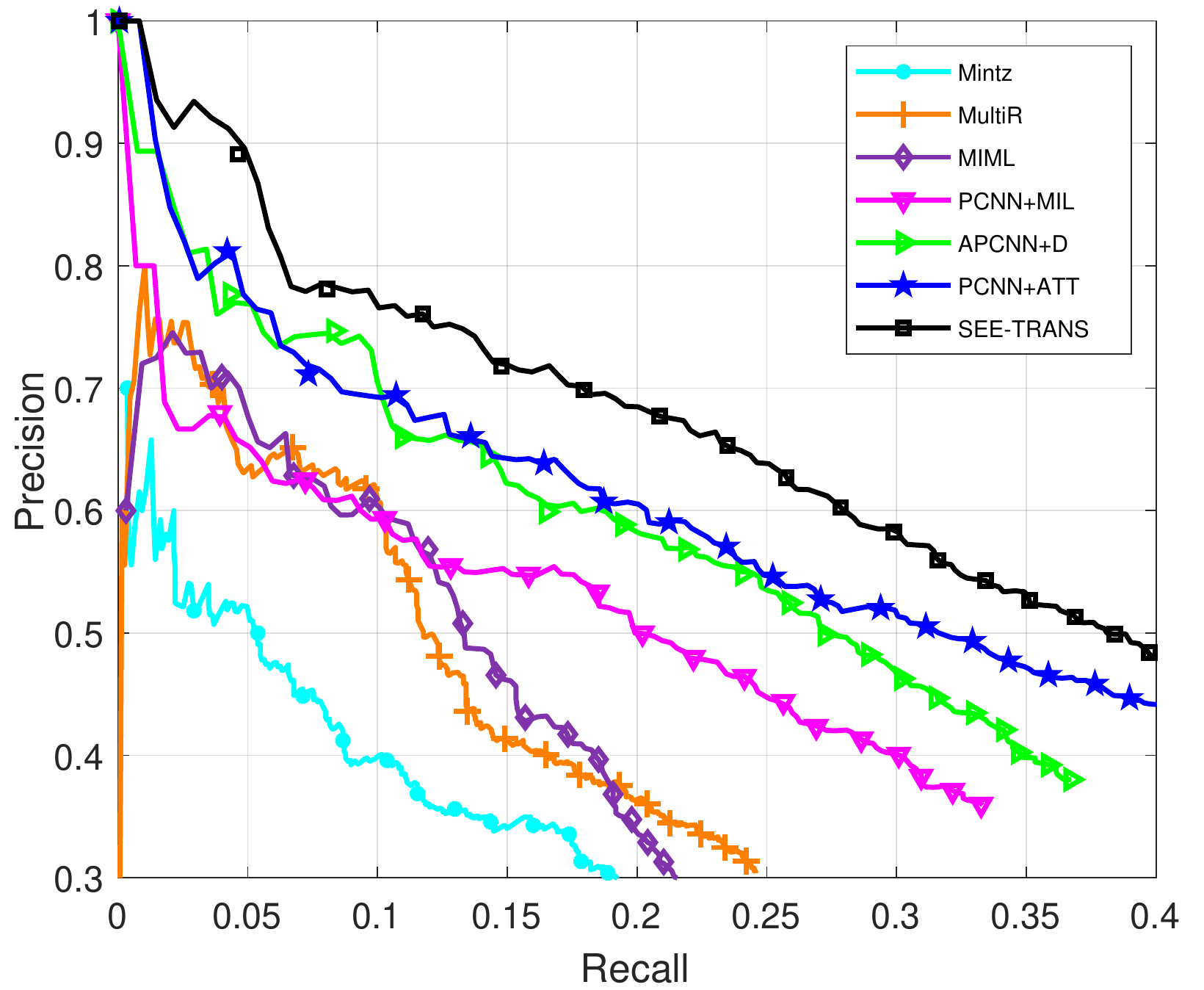}
      \end{minipage}
  \caption{Comparison with previous results.}\label{heldout}
\end{figure}

\textbf{Comparison with previous works} is presented in  Figure \ref{heldout}.
We select six representative approaches and directly get all their results from  \citeauthor{Lin2016Neural} \shortcite{Lin2016Neural} and \citeauthor{ji2017distant} \shortcite{ji2017distant} for comparison\footnote{We are very grateful to Dr. Lin and Dr. Ji for their help.}, which fall into two categories:
\begin{itemize}

\item Traditional discrete feature-based methods: (1) \textbf{Mintz} \cite{mintz2009distant} proposes distant supervision paradigm and uses a multi-class logistic regression for classification. (2) \textbf{MultiR} \cite{hoffmann2011knowledge} is a probabilistic graphical model with multi-instance learning under the ``at-least-one" assumption. 
    (3) \textbf{MIML} \cite{surdeanu2012multi} is also a graphical model with both multi-instance and multi-label learning.

\item Neural model-based methods: (1) \textbf{PCNN+MIL} \cite{zeng2015distant} proposes piece-wise (three-segment) CNN to obtain sentence embeddings. (2) \textbf{PCNN+ATT} \cite{Lin2016Neural} corresponds to our baseline approach and achieves state-of-the-art results. (3) \textbf{APCNN+D} \cite{ji2017distant} uses external background information of entities via an attention layer to help relation classification.
\end{itemize}

From the results, we can see that our proposed approach ``SEE-TRANS'' consistently outperforms all other approaches by large margin, and achieves new state-of-the-art results on this dataset, demonstrating the
effectiveness of leveraging syntactic context for better entity representation for distant supervision relation extraction.

\subsection{Manual Evaluation}

Due to existence of noises resulting from distance supervision in the test dataset under the held-out evaluation, 
we can see that there is a sharp decline in the precision-recall curves in most models 
in Figure \ref{heldout}.
Therefore, we manually check the top-$500$ entity-relation tuples returned by all the eight approaches.\footnote{Please note that there are many overlapping results among different approaches, thus requiring much less manual effort.}
Table \ref{manual} shows the results. 
We can see that (1) our re-implemented baseline achieve nearly the same performance with \citeauthor{Lin2016Neural} \shortcite{Lin2016Neural}; (2) our proposed SEE-TRANS achieves consistently higher precision at different $N$ levels. 

\begin{table}[tb]
  \centering
  \small
  \begin{tabular}{c|cccc}
    \hline
    Accuracy & Top 100 & Top 200 & Top 500 & Average \\
    \hline
    Mintz    & 0.77 &0.71 & 0.55 & 0.676 \\
    MultiR   & 0.83 &0.74 & 0.59 & 0.720 \\
    MIML     & 0.85 &0.75 & 0.61 & 0.737 \\
    PCNN+MIL & 0.84 &0.77 & 0.64 & 0.750 \\
    PCNN+ATT & 0.86 &0.83 & 0.73 & 0.807 \\
    APCNN+D  & 0.87 &0.83 & 0.74 & 0.813 \\
    \hline
    Baseline & 0.86 & 0.84 & 0.73 & 0.810 \\
    SEE-TRANS & \textbf{0.91} & \textbf{0.87} & \textbf{0.77 }& \textbf{0.850} \\
    \hline
  \end{tabular}
  \caption{Manual evaluation results.}\label{manual}
\end{table}

\begin{table*}[tb]
\centering
\small
\begin{tabular}{|c|  m{0.38\textwidth}   |   c |   c   |  c|c|}
  \hline
  \textbf{Tuple} & \textbf{Sentences} & \textbf{Syntax-aware Entities} & \textbf{Baseline} & \textbf{SEE-TRANS} \\
  \hline
  \multirow{4}[20]{0.10\textwidth }{\emph{company } (Bruce Wasserstein, Lazard)}
  & 1. A record profit at [Lazard], the investment bank run by [Bruce Wasserstein], said that strength in its merger advisory ...
  & \multirow{4}[25]{0.15\textwidth}{\textbf{Bruce Wasserstein}: \\ 1. the chairman of Lazard.  \\ 2. the current Lazard chief executive.
  \newline \newline \textbf{Lazard}: \\ 1. the investment bank run by Bruce Wasserstein.  }
  & \multirow{4}[25]{0.1\textwidth}{\emph{NA} (0.735)\newline\newline \emph{company} (0.256) \newline\newline \emph{founders} (0.002)}
  & \multirow{4}[25]{0.1\textwidth}{\emph{company} (0.650) \newline\newline \emph{NA} (0.250) \newline\newline \emph{founders} (0.028)} \\
  \cline{2-2}

  & 2. The buyout executives ... huddled in a corner, and [Bruce Wasserstein], the chairman of [Lazard], chatted with richard d. parsons , the chief executive of time warner .
  &
  &   &   \\
  \cline{2-2}

  & 3. [Lazard], the investment bank run by [Bruce Wsserstein], said yesterday that strength in its merger-advisory ...
  &
  &   &  \\
  \cline{2-2}

  & 4. Along with the deals and intrigue ...  maneuverings in martha 's vineyard as well as the tax strategies of the current [Lazard] chief executive [Bruce Wasserstein].
  &
  &   &   \\
  \hline

\end{tabular}

  \caption{Case study: a real example for comparison. }\label{case_study}
\end{table*}

\subsection{Case Study}

Table \ref{case_study} present a real example for case study. The entity-relation tuple is (\emph{Bruce Wasserstein}, \emph{company}, \emph{Lazard}). There are four sentences containing the entity pair.
The baseline approach only uses the word sequences as the input, and learn the sentence embeddings for relation classification.
Due to the lack of sufficient information, the \emph{NA} relation type receives the highest probability of $0.735$.
In contrast, our proposed SEE-TRANS can correctly recognize the relation type as \emph{company} with the help of the rich contexts in the syntactic parse trees.

\section{Related Work}
In this section, we first briefly review the early previous studies on distant supervision for RE. Then we introduce the systems using the neural RE framework.

In the supervised paradigm, relation extraction is considered to be a multi-class classification problem and needs a great deal of annotated data, which is time consuming and labor intensive. To address this issue, \citeauthor{mintz2009distant} \shortcite{mintz2009distant} aligns plain text with Freebase by distant supervision, and extracts features from all sentences and then feeds them into a classifier. However, the distant supervision assumption neglects the data noise.
To alleviate the wrong label problem, \citeauthor{riedel2010modeling} \shortcite{riedel2010modeling} models distant supervision for relation extraction as a multi-instance single-label problem. Further, \citeauthor{hoffmann2011knowledge} \shortcite{hoffmann2011knowledge} and \citeauthor{surdeanu2012multi} \shortcite{surdeanu2012multi} adopt multi-instance multi-label learning in relation extraction, and use the shortest dependency path as syntax features of relation. The main drawback of these methods is that their performance heavily relies on a manually designed set of feature templates which are difficult to design.

Neural networks  \cite{bengio2009learning} have been successfully used in many NLP tasks such as part-of-speech tagging \cite{santos2014learning}, parsing \cite{socher2013parsing}, sentiment analysis \cite{dos2014deep}, and machine translation \cite{cho-al-emnlp14}.
As for relation extraction, neural networks have also been successfully applied and achieved advanced performance for this field.
\citeauthor{Socher2012Semantic} \shortcite{Socher2012Semantic} uses a recursive neural network in relation extraction.
\citeauthor{Zeng2014Relation} \shortcite{Zeng2014Relation} adopts an end-to-end convolutional neural network in this task, and \citeauthor{zeng2015distant} \shortcite{zeng2015distant} further combines at-least-one multi-instance learning and assumes that only one sentence expresses the relation for each entity pair, which doesn't make full use of the supervision information.
\citeauthor{Lin2016Neural} \shortcite{Lin2016Neural} proposes to use attention to select valid sentences, which shows promising results. However, sentence embeddings are used to represent relation between entities, may result in  semantic shifting problem, since the relation between entities is just a small part of a sentence.

All the above work on neural networks mainly use words to generate sentence embeddings, and use them for classification. Besides the word-level information, syntax information also has been considered by some researchers, for example,
\citeauthor{miwa2016endtoend} \shortcite{miwa2016endtoend} and \citeauthor{cai2016bidirectional} \shortcite{cai2016bidirectional} 
model the shortest dependency path as a factor for the relation between entities, but they ignore that the tree information can be used to model the syntax roles the entities played. The syntax roles are important for relation extraction.
Different from the above previous studies, we enrich the entity representations with syntax structures by considering the subtrees rooted at entities.

\section{Conclusion}

In this paper, we propose to learn syntax-aware entity embedding from dependency trees for enhancing
neural relation extraction under the distant supervision scenario.
We apply the recursive tree-GRU to learn sentence-level entity embedding in a parse tree, and utilize both intra-sentence and inter-sentence attentions to make full use of syntactic contexts in all sentences.  We conduct
experiments on a widely used benchmark dataset. The experimental results show that our model consistently outperforms
both the baseline and the state-of-the-art results.
This demonstrates that our approach can effectively learn entity embeddings, and the learned embeddings are able to help the task of relation extraction.

For future, we would like to further explore external knowledge as  \citeauthor{ji2017distant} \shortcite{ji2017distant} to obtain even better entity embeddings.
We also plan to apply the proposed approach to other datasets or languages.

\section{ Acknowledgments}
The research work is supported by the National Key Research and Development Program of China under Grant No.2017YFB1002104,
and the National Natural Science Foundation of China (61672211). This work is partially supported by the joint research project of Alibaba and Soochow University.

\bibliographystyle{aaai}
\bibliography{references}

\end{document}